# Title: Prediction of Alzheimer's Disease Risk Factors from Retinal Images via Deep Learning: Development and Validation of Biologically Relevant Morphological Associations in the UK Biobank

Author: Seowung Leem[1], Yunchao Yang Ph.D.[2,3], Adam J. Woods Ph.D. [4], Ruogu Fang Ph.D.[1,5,6,7*]

[1]J. Crayton Pruitt Family Dept. of Biomedical Engineering, University of Florida, Gainesville, FL 32611, USA

[2] University of Florida Research Computing, University of Florida, Gainesville, FL 32611, USA

[3] Meta AI (FAIR)

[4] School of Behavioral and Brain Sciences, University of Texas at Dallas, Richardson, TX. 75080, USA

[5] Dept. of Electrical and Computer Engineering, University of Florida, Gainesville, FL 32611, USA

[6] Dept. of Computer and Information Science and Engineering, University of Florida, Gainesville, FL. 32611, USA

[7] Center for Cognitive Aging and Memory, University of Florida, Gainesville, FL 32611, USA

*Corresponding Author:

Ruogu Fang, Ph.D.

J. Crayton Pruitt Family Department of Biomedical Engineering

Herbert Wertheim College of Engineering

University of Florida

PO Box 116131, 1275 Center Drive

Gainesville, FL 32611-6131

Phone:(352) 294 1375

Email: ruogu.fang@bme.ufl.edu

## Abstract

*Background*

The systemic, metabolic, lifestyle factors have established associations with Alzheimer's Disease (AD) through epidemiologic and AD-specific biomarker studies. Whether colored fundus photography (CFP) contains retinal structural signatures corresponding to these AD-related risk domains remains unclear.

*Objective*

To determine whether deep learning (DL) models can predict 12 AD-related risk factors from CFP and to characterize the retinal structures underlying these predictions, thereby assessing whether CFP reflects pathways to AD vulnerability.

*Methods*

Using 62,876 CFPs from 44,501 unique participants from the UK Biobank , DL models were trained to predict 12 factors linked to AD pathology or incidence: 6 categorical (sex, smoking, sleeplessness, economic status, alcohol use, depression) and 6 continuous (age, age at completing education, BMI, systolic, diastolic blood pressure, HbA1c). Model performance, model saliency, and saliency-derived scores (CAM-Score) were evaluated and compared to retinal morphometry. The scores were also compared between incident-AD cases (average 8.55 years before onset) and matched controls.

*Results*

Predictive performance of DL ranged from AUROC between 0.5654 and 0.9480 for categorical factors and $R^2$ between -0.0291 and 0.7620 for continuous factors, outperforming most of the morphometry-based machine learning models. Saliency-based score consistently highlighted biologically meaningful regions, particularly the optic nerve head and retinal vasculature. It also aligned with present morphometric variations. Several saliency-based scores differed significantly between incident AD and matched controls, suggesting potential overlap between retinal correlates of AD-related risk factors and preclinical AD-associated changes.

*Conclusion*

CFP encodes retinal signatures linked to AD risk factors. Although not diagnostic, DL-derived retinal representations may uncover biologically meaningful risk-related structural changes mirroring the potential AD vulnerability.

## Introduction

Alzheimer's Disease (AD) is a leading cause of dementia, affecting 6.9 million in the United States and over 50 million worldwide [1,2]. In the absence of curative therapies, prevention and risk reduction strategies remain central to mitigating disease burden. Increasing evidence indicates that AD is a multifactorial disorder influenced by cardiovascular, metabolic, and lifestyle factors, including hypertension, diabetes, obesity, and smoking [3–8]. Accordingly, identifying individuals at elevated risk through scalable and non-invasive approaches is critical for enabling early intervention prior to clinical symptom onset

Current approaches to AD risk assessment rely on combinations of cognitive testing [9–11], clinical evaluation [12–14], and fluid biomarkers such as amyloid-β and phosphorylated tau [15–18]. While these methods provide valuable insights, they are often resource-intensive, invasive, or not readily accessible for large-scale screening. In contrast, retinal imaging offers a practical and non-invasive alternative for population-level risk stratification. The retina reflects systemic vascular and metabolic health, and color fundus photography (CFP) is widely available, cost-effective, and routinely used in clinical care [19–21].

Recent advances in deep learning (DL) have enabled the extraction of systemic health information from retinal images beyond traditional ophthalmic assessment. Prior studies have demonstrated that CFP can be used to predict demographic and clinical variables such as age, blood pressure, glycemic status, and smoking history with high accuracy [22–24]. These factors are well-established contributors to AD risk [6,7], suggesting that retinal images may encode composite signatures of systemic health relevant to neurodegenerative vulnerability. In contrast, electronic health record (EHR)-derived measurements are typically sparse and episodic, limiting their ability to capture the cumulative biological impact of these exposures over time. Retinal morphology, by reflecting microvascular and neurovascular integrity, may therefore provide an integrated representation of long-term risk. However, existing work has largely focused on predicting individual risk factors or disease states in isolation, without systematically examining how retinal structural patterns relate to a broader spectrum of AD-related risk domains.

A critical gap therefore remains in understanding whether retinal features captured by CFP can serve as proxies for upstream risk factors that contribute to AD development. Such an approach shifts the focus from detecting disease pathology to modeling risk profiles, aligning more closely with preventive strategies and enabling scalable screening. Furthermore, while CFP is the most accessible retinal imaging modality, its potential for comprehensive risk factor profiling remains underexplored compared to more specialized imaging techniques [19–21].

In this study, we propose that CFP contains structural signatures associated with systemic and lifestyle risk factors relevant to AD. We develop an integrative framework combining DL-based prediction, explainability analysis, and statistical validation to characterize these associations. Using the UK Biobank, we train models to predict 12 AD-related risk factors from retinal images. To interpret model behavior, we introduce a saliency-based approach to localize retinal regions contributing to each prediction and evaluate their biological plausibility through comparison with quantitative retinal morphology. Finally, we examine whether these risk-related retinal signatures are associated with future incident AD, providing insight into their relevance for risk stratification.

## Methods

*Study Overview*

This study consists of two components. The primary analysis was designed as a cross-sectional modeling evaluating whether retinal images obtained from the CFP contain structural signatures associated with systemic risks related to AD. Two deep learning (DL) models were developed to predict Alzheimer's disease–related demographic, vascular, metabolic, and lifestyle risk factors using retinal images and corresponding risk factor data from the UK Biobank. The image and risk factor were collected in the same visit, assuring the cross-sectional design of the study.

In addition to this primary cross-sectional analysis, we conducted an additional exploratory retrospective study to examine whether retinal features associated with these risk factors differed between individuals who later developed incident AD and demographic-matched controls. In this section, baseline retinal images obtained at the initial assessment were analyzed, and participants were subsequently classified as having incident AD during the follow-up. Importantly, this analysis is exploratory and based on the retrospective outcome. The result should not be interpreted as evidence of predictive modeling for incident AD.

*Study Population*

This study utilized CFP images and associated risk factors sourced from the UK Biobank [25]. UK Biobank is a large-scale prospective cohort study comprising over 500,000 participants aged 40 to 69 years. The baseline assessment data were collected between 2006 and 2010, and the analyses were restricted to baseline assessments only.

*Dataset*

From the UK Biobank, twelve AD-associated risk factors were selected based on the extensive prior studies investigating the risk factors based on the context of dementia [3,8,26] and specific AD-related biomarkers [4–7,27–29]. To leverage the strengths of CFP in noninvasiveness and accessibility, the primary focus of this study was placed on modifiable risk factors related to lifestyle, cardiovascular health, and metabolic function. The selected variables included genetic sex, sleeplessness/insomnia, smoking status, alcohol intake frequency, bipolar and major depression status, average total household income before tax, age at assessment center attendance, age at completion of full-time education, body mass index, diastolic blood pressure, systolic blood pressure, and glycated hemoglobin.

Data were obtained through the UK Biobank's publicly available assessment protocols (https://www.ukbiobank.ac.uk/). Questionnaire-based variables (e.g., sleeplessness/insomnia, smoking status, alcohol intake frequency, bipolar and major depression status, average total household income before tax, age at assessment center attendance, age at completion of full-time education) were self-reported using UK Biobank assessment center touchscreen interfaces. Physical measures (body mass index and automated readings of systolic and diastolic blood pressure) were collected via standardized procedures at the assessment center. HbA1c levels and genetic sex were derived from biospecimens and processed at UK Biobank laboratories following the initial assessment.

For retinal images, approximately 67,000 subjects underwent the imaging session using Topcon 3D OCT-1000 MK2 (https://www.ukbiobank.ac.uk/enable-your-research/about-our-data/imaging-data). The collected CFP consists of left (n=68,177) and right images (n=68,767). The primary field of view (FOV) was 45°, and the pixel resolution (mm per pixel) was not available from the image metadata.

*Image Selection, Risk Factor Harmonization, and Dataset Partitioning*

Participants from the cohort with at least one CFP from either eye were included. Image quality was assessed using an automated DL based quality grading tool for fundus images [30]. The CFPs were categorized as good, usable, or poor based on the given probability of the grading tool.

CFPs graded as good or usable were retained for the analysis. After image quality check, 62,876 images from 44,501 unique participants were retained. To standardize the orientation of retinal structures, right-eye images were horizontally flipped to align with left-eye anatomy.

Among retained participants, 12 variables known to influence AD risk were collected, comprising 6 categorical and 6 continuous variables. To optimize the model's discriminative performance, four categorical variables (smoking status, alcohol intake frequency, bipolar and major depression, and average total house income before tax) were reclassified. For the continuous variable, blood pressure measures (DBP and SBP), which were taken a few moments apart, were averaged as a single value.

After all preprocessing, participants were randomly assigned (stratified by risk factor distributions to handle the imbalance between data splits) to a development set (80%) and a validation set (20%). Missing values due to non-responses were handled using a two-stage imputation pipeline. For categorical factors, missing values were imputed using a histogram-based gradient boosting classifier, where the missing variable was modeled as a function of the remaining risk factors [31]. Continuous factors were imputed via k-nearest neighbors (k=5) with missing values replaced by the mean of the five nearest neighbors [32]. Imputation models were trained exclusively on the development set to prevent data leakage. Missing data in the validation set were excluded during analysis. For the explainability analysis, 1152 subjects with risk factors and no missing data from the validation set were included.

In our analysis, despite the data split of development and validation set being performed based on the subject level, each eye was treated as an independent observation in both model development and statistical analyses, as previous studies [22,23]. When data from both eyes of a participant were available, both were included without explicit modeling. This approach was adopted because the primary analysis of investigating the relationship between the individual eye and risk factors was formulated at the eye level. Subject-level clustering or eye selection was therefore not applied.

For clarity, original UK Biobank variable labels were standardized for presentation throughout the manuscript: Genetic sex (Sex), Sleeplessness/insomnia (Sleeplessness), Smoking status (Smoking), Alcohol intake frequency (Alcohol Use), Bipolar and major depression status (Recurrent Depression Status), Average total household income before tax (Economic Status), Age at assessment center attendance (Age), Age at completion of full-time education (Education), Body mass index (BMI), Diastolic blood pressure (DBP), Systolic blood pressure (SBP), and Glycated hemoglobin (HbA1c). Summary statistics for both development and validation cohorts are detailed in Table 1.

**Table 1. Summary Statistics of Patients' Risk Factors in the Development and Validation Sets.**

| Risk Factors | UK Biobank | | | | P-value | Effect Size | Citation |
|---|---|---|---|---|---|---|---|
| | Development set (n=35,416) | | Validation set (n=9,085) | | | | |
| | Categorical; | Continuous | Categorical | Continuous | | | |
| Sex: male % | 45.32 | n/a | 45.10 | n/a | 0.96 | 0.00 | [6] |
| Age: mean years (s.d.) | n/a | 55.51 (8.22) | n/a | 55.73 (8.23) | 0.04 | 0.02 | [6,7] |
| Education: mean years (s.d.) | n/a | 17.00 (2.41) | n/a | 16.97 (2.38) | 0.35 | 0.01 | [29] |
| Sleeplessness: % | 26.25 Never/Rarely 47.25 Sometimes 26.49 Usually | n/a | 26.27 Never/Rarely 47.19 Sometimes 25.91 Usually | n/a | 0.96 | 0.01 | [4,5,27] |
| Smoking: positive % | 43.34 | n/a | 42.57 | n/a | 0.75 | 0.01 | [6,7] |
| Alcohol Use: % | 19.58 Low 36.73 Moderate 43.69 Excessive | n/a | 20.48 Low 36.88 Moderate 42.63 Excessive | n/a | 0.85 | 0.01 | [6] |
| Recurrent Depression Status: % | 26.37 | n/a | 26.53 | n/a | 0.96 | 0.00 | [7] |
| Economic Status: % | 18.92 Low 49.57 Middle | n/a | 19.28 Low 50.48 Middle | n/a | 0.83 | 0.01 | [28] |

| | 31.50 High | | 30.24 High | | | | |
|---|---|---|---|---|---|---|---|
| BMI: mean (s.d.) | n/a | 27.21 (4.72) | n/a | 27.22 (4.78) | 0.82 | 0.00 | 6 |
| DBP: mean mmHg (s.d.) | n/a | 81.90 (10.74) | n/a | 81.79 (10.61) | 0.39 | 0.01 | 7 |
| SBP: mean mmHg (s.d.) | n/a | 138.91 (19.56) | n/a | 138.72 (19.25) | 0.41 | 0.01 | 7 |
| HbA1c: mean mmol/mol (s.d.) | n/a | 35.75 (6.43) | n/a | 35.77 (6.33) | 0.75 | 0.00 | 7 |

* The p-value for categorical variables is determined using the Chi-squared test, while for continuous variables, it is calculated using the independent t-test. The effect size for categorical variables was computed by Cramer's v, while Cohen's d was computed for continuous variables.

*Deep Learning Model Architecture and Training*

The DL framework in this study was based on the Shifted Window (Swin) Vision Transformer [33]. The Swin Transformer is a self-attention-based architecture, utilizing shifted windows to compute a hierarchical visual representation for computer vision tasks. Transfer learning was used to initialize the model with pretrained weights from the ImageNet dataset [34], allowing the network to benefit from priors from the large scale natural image data to accelerate the optimization process. The only modification in the architecture was applied on the final classification layer to accommodate the model for the AD risk factor prediction task. Given the distinction between categorical and continuous risk factors, two separate models were trained: for classification and regression, each. The classification model was optimized by minimizing the categorical cross-entropy loss with weights derived from the development set to consider class imbalance, while the regression model was optimized by minimizing the mean squared error between ground truth and the model's prediction. For both models, individual losses for risk factors were calculated separately and aggregated into a single loss function to update the model parameters. Model training was conducted using the AdamW optimizer with a learning rate of 1e4. The total number of training epochs was 100, and the parameters yielding the best performance on the development set were selected for evaluation on the validation set. In this study, high performance computational resources provided by the University of Florida's

HiperGator AI system were used, employing 32 NVIDIA DGX A100 GPUs to expedite the training process.

*Model Evaluation*

DL model performance was assessed separately for classification and regression tasks based on the type of each risk factor (Table 1). For categorical outcomes, evaluation metrics included balanced classification accuracy and the area under the receiver operating characteristic curve (AUROC). For continuous outcomes, the coefficient of determination ($R^2$) was used to quantify the model fit. A nonparametric bootstrapping procedure was applied to assess metric stability. Random resampling with replacement from the validation set generated empirical distributions of each metric, from which 95% confidence intervals (CI) were derived.

To establish baseline comparisons and assess whether DL models outperform simpler approaches, 2 different morphometry-based machine learning (ML) models were trained and evaluated for each risk factor. Specifically, linear (or logistic) regression (LR) and histogram-based gradient boosting (HGB) models were applied to retinal morphometric features to predict both categorical and continuous risk factors. For input features, 24 retinal morphometry values from CFP were leveraged (see "*Validation of DL-Derived Features Biological Relevance Using Retinal Morphology Metrics"* section for details about morphometry features). The missing value was imputed with the mean values of each feature in the training set distribution. Models were optimized using grid search on the development set, and the best-performing configuration was subsequently evaluated on the validation set (grid search parameters and best-performing configurations are available in Supplementary Table 1).

*Model Interpretability and Saliency Analysis*

While saliency methods alone do not guarantee mechanistic interpretability, quantification of population-level overlap between model saliency and anatomically defined structures provides an empirical test of whether the model's attention is concentrated in retinal regions known to reflect systemic changes. Prior studies frequently rely on qualitative inspection of Grad-CAM heatmaps, which is subjective and difficult to reproduce [22,24]. For the DL model's interpretability, class activation mapping (CAM) was used to generate saliency maps highlighting the retinal regions contributing most to the prediction of each risk factor. Specifically, the Score-CAM method [35], a gradient-free approach known for producing sharper and more interpretable saliency maps than gradient-based approaches, was used. The first normalization layer in the last Swin Transformer block was chosen as a target layer for saliency map generation. While saliency maps provide visual insights into a model's prediction at the individual level, their interpretability is often subjective and limited by observer expertise. To address this limitation, the CAM-Score is introduced in this study to evaluate the model's attention to anatomically defined retinal structures at the population level. For a given retinal structure (k), the CAM-Score ($r^k.$) was calculated using Equation 1. In Equation 1, binary segmentation makes ($M_{ij}^k$) of four major retinal structures visible from CFP, artery, vein, optic cup, and disc ($k$), was generated. Thresholding was applied with threshold ($t$) to binarize the saliency map ($S_{ij}^k$). $1(\cdot)$ is the indicator function used to threshold the saliency map at a value t to binarize it. The overlap ratio between the retinal structure segmentation and the binarized saliency map was then computed by dividing the sum of overlapped pixels by the total number of pixels in the segmentation map. Thus, the CAM-Score reflects the proportion of pixels within a given anatomical structure that overlap with the most salient regions identified by the DL model. The CAM-Score formalizes the evaluation of a model's interpretability by providing a structure-specific attribution metric. Although it is not causal, the CAM-Score enables stronger validity than visual inspection alone and allows downstream correlation with independent morphometric measurements.

$$r^k = \sum_{ij} \frac{M_{ij}^k \cdot 1(S_{ij}^k \geq t)}{M_{ij}^k} \quad (1)$$

*Validation of DL-Derived Features Biological Relevance Using Retinal Morphology Metrics*

To assess whether the DL model infers retinal features associated with AD risk during prediction, two complementary analyses were conducted. First, linear models were trained to predict established retinal morphology measures from DL-derived feature embeddings of CFP. These embeddings, extracted from the same target layer from Swin Transformer for saliency mapping, represent each fundus image as a 1,536-dimensional vector. A separate linear model was trained for each retinal morphological feature (n=24) from 2 different zones: Zone B (0.5-1.0 optic disc diameter) and Zone C (0.5-2.0 optic disc diameter), using both classification and regression models. This yielded 96 linear models, and each was fitted and validated with 10-fold cross-validation. Second, the relationship between risk factors and retinal morphology features was analyzed to find a significant relationship and how they map to our model's inference on retinal structures. The non-parametric permutation test with 10,000 random permutations was applied to analyze the relationship. The independent two-sample t-test was used for categorical risk factors to compare the distributions of morphology features among risk factor groups. Pearson's correlation coefficient (R) was used for continuous risk factors to quantify linear associations with retinal morphology. For both tests, statistical significance was determined by P<.05.

24 retinal morphology features were examined in this analysis. These included the width and height of optic cup and disc, vertical and horizontal cup-to-disc ratios (CDR), fractal dimension [36], vessel density, arterial and venous tortuosity (distance tortuosity, squared curvature tortuosity [37], and tortuosity density [38]), and vessel calibers (Central Retinal Arterial Equivalent (CRAE), Central Retinal Venular Equivalent (CRVE), Arteriole-to-venule ratio (AVR)) assessed using the Par-Hubbard [39] and Knudtson [40] methods. The descriptions of metrics, which are derivative, not direct measures, are summarized in Table 2.

Morphological features were computed in two concentric annular regions centered on the optic disc: Zone B (0.5-1.0 optic disc diameter) and Zone C (0.5-2.0 optic disc diameter). Additionally, optic disc and cup dimensions were measured in pixels (912x912), as absolute physical scale calibration was not available from the metadata of image files from the UK Biobank.

*Retrospective analysis of CAM scores between the normal and incident Alzheimer's Disease population*

As an exploratory analysis, we compared saliency derived retinal features between UK Biobank participants who after developed incident AD during follow-up and matched controls without AD. To evaluate whether the DL model-derived features from the risk factor prediction network reflect meaningful retinal structural differences between the normal population and those who later developed Alzheimer's Disease, we analyzed CAM-scores across four retinal structures. Incident AD cases were identified using the UK Biobank report dates for three different dementia related classifications: AD (Field ID: 42020), Vascular dementia (VD) (Field ID: 42022), and frontotemporal dementia (FTD) (Field ID: 42024). We excluded all participants with VD or FTD to retain only individuals with incident AD without comorbid dementia subtypes (mean=8.55 years, range=2.38-11.41). A total of 52 subjects met the inclusion criteria. For comparison, we selected 52 dementia-free subjects matched using stratified sampling to ensure no significant differences in key risk factors between groups (details available in Supplementary Table 2). CAM-Scores were compared between the incident AD and matched normal cohort using the Mann-Whitney U test. Subjects with non-zero CAM-Scores were only included in the analysis.

**Table 2. Summary of derived retinal morphological features and their associations with systemic risks.**

| Derived Retinal Morphology Features | Summary | Findings in previous studies* |
|---|---|---|
| Fractal Dimension | A numerical value quantifying the complexity. A higher value indicates a more developed complex branching of the retinal vasculature. | Patients with low fractal dimension had stroke and cerebrovascular disease [41]. |
| Vessel Density | A numerical value quantifying the image area of CFP covered by the vessel network. | Patients with cerebrovascular damage and CSF dysfunction had reduced vessel density [42]. |
| Distance Tortuosity | A metric quantifying how much vessels bent and twisted by calculating the ratio between the vessel path length and the straight line distance between endpoints. | Patients with vascular abnormalities and hypertension had higher distance tortuosity [43]. |
| Squared Curvature Tortuosity | A computational metric quantifying local vessel bending by dividing the square of the curvature along the vessel path by the total length. It is more sensitive to subtle changes in vessel shape. | Elevated squared curvature torruosity in the artery was observed in patients with older age, higher blood pressure, and higher BMI, while decreased value in the vein was observed in patients with younger age, higher blood pressure, and lower high-density lipoprotein (HDL) [44]. |
| Tortuosity Density | A computational metric quantifying the prevalence and distribution of curly segments across the retina by combining tortuosity severity and frequency. | Patients with ocular and systemic diseases [45] had higher tortuosity density. |
| CRAE & CRVE | Computational metrics compute the average diameter of the six largest arteries or veins within a standardized zone around the optic disc. | Patients with hypertension and cardiovascular risk had narrower CRAE and wider CRVE [46]. |
| AVR | The ratio computed by CRAE/CRVE. | Patients with increased cardiovascular risk had lower AVR [46]. |

* Findings of previous reported relationships. They are not conclusive.

## Results

**Table 3. Model Prediction Result of Alzheimer's Disease Risk Factor.**

| Risk Factors | Prediction Result | | |
|---|---|---|---|
| | Metric | Performance (95% CI)* | Chance Probability |
| Sex | Balanced Accuracy | 0.8677 (0.8650, 0.8703) | 0.5 |
| | AUROC | 0.9480 (0.9464, 0.9496) | 0.5 |
| Age | $R^2$ | 0.7620 (0.7583, 0.7656) | 0 |
| Education | $R^2$ | -0.0291 (-0.0363, -0.0222) | 0 |
| Sleeplessness | Balanced Accuracy | 0.3963 (0.3924, 0.4002) | 0.3333 |
| | AUROC | 0.5654 (0.5621, 0.5687) | 0.5 |
| Smoking | Balanced Accuracy | 0.5751 (0.5714 0.5790) | 0.5 |
| | AUROC | 0.6145 (0.6098, 0.6193) | 0.5 |
| Alcohol Use | Balanced Accuracy | 0.4374 (0.4309, 0.4438) | 0.3333 |
| | AUROC | 0.6127 (0.6075, 0.6178) | 0.5 |
| Recurrent Depression Status | Balanced Accuracy | 0.5583 (0.5520, 0.5645) | 0.5 |
| | AUROC | 0.5858 (0.5781 0.5935) | 0.5 |
| Economic Status | Balanced Accuracy | 0.4709 (0.4649, 0.4770) | 0.3333 |
| | AUROC | 0.6417 (0.6366, 0.6469) | 0.5 |
| BMI | $R^2$ | 0.0994 (0.0935, 0.1055) | 0 |
| DBP | $R^2$ | 0.2725 (0.2653, 0.2795) | 0 |
| SBP | $R^2$ | 0.3281 (0.3211, 0.3347) | 0 |
| HbA1c | $R^2$ | 0.0926 (0.0846, 0.1003) | 0 |

*Bootstraping CIs based on the 2000 resamplings for each metric.

*DL Model Captures Systemic Changes Related to AD Risk in Retina.*

The predictive performance of our model across the 12 AD risk factors varied (Table 3). For categorical risk factors, the strongest performance was observed in predicting sex. An accuracy of 0.8677 (95% CI: 0.8650-0.8703) and an AUROC of 0.9480 (95% CI: 0.9464-0.9496) indicate the excellent discriminative performance exceeding random chance level. The above chance but limited performance was observed in smoking (accuracy: 0.5751; 95% CI, 0.5714-0.5790; AUROC: 0.6145; 95% CI, 0.6098-0.6193) and alcohol use (accuracy: 0.4374; 95% CI, 0.4309-0.4438; AUROC: 0.6127; 95% CI, 0.6075-0.6178), indicating non-random association between retinal features and these factors. For economic status, depression and sleeplessness, the accuracy of 0.4709 (95% CI, 0.4649-0.4770), 0.5583 (95% CI, 0.5520-0.5645), and 0.3963 (95% CI, 0.3924-0.4002), and AUROC values of 0.6417 (95% CI, 0.6366-0.6469), 0.5858 (0.5781-0.5935) and 0.5654 (95% CI, 0.5621-0.5687) was lower, indicating weak associations that may reflect limited or indirect relationships between retinal morphology and AD-related risk factors.

For continuous risk factors, the robust predictive performance was observed for age with an $R^2$

of 0.7620 (95% CI, 0.7583-0.7656). This suggests that the age-related retinal structural differences were well captured by the DL model. SBP and DBP were predicted with limited performance above chance with $R^2$ of 0.3281 (95% CI, 0.3211-0.3347) and 0.2725 (95% CI, 0.2653-0.2795), respectively. The observed performance in BMI ($R^2$: 0.0994; 95% CI, 0.0935-0.1055) and HbA1c ($R^2$: 0.0926; 95% CI, 0.0846-0.1003) was weak with a limited predictive signal. However, the prediction of Education was in a negative $R^2$ (−0.0291; 95% CI, −0.0363 to −0.0222), meaning no retinal correlation.

Relative to simpler models based on CFP-derived morphometric features (Supplementary Table 3 for categorical risk factors, Supplementary Table 4 for continuous risk factors), DL models achieved superior performance across most risk factor prediction tasks. For several key risk factors, including sex, smoking status, alcohol use, economic status, age, and SBP and DBP, the DL models consistently outperformed the simpler morphometry-based ML approaches. In contrast, modest performance gains were observed for sleeplessness, recurrent depression status, BMI, and HbA1c, whereas no improvement was observed for education.

**Table 4. CAM-Scores Across Retinal Regions by Categorical Risk Factor Groups.**

| **Risk Factors** | **Class** | **CAM-Score; mean(std)** | | | |
|---|---|---|---|---|---|
| | | **Artery** | **Vein** | **Optic Disc** | **Optic Cup** |
| Sex | Female | 0.10(0.07) | 0.09(0.07) | 0.02(0.11) | 0.01(0.09) |
| | Male | 0.14(0.09) | 0.16(0.09) | 0.68(0.43) | 0.70(0.38) |
| Smoking | Never Smoked | 0.08(0.06) | 0.07(0.06 | 0.06(0.22) | 0.06(0.19) |
| | Smoked | 0.08(0.06) | 0.09(0.06) | 0.73(0.40) | 0.66(0.36) |
| Sleeplessness | Never | 0.09(0.08) | 0.09(0.08) | 0.44(0.45) | 0.45(0.39) |
| | Sometimes | 0.04(0.04) | 0.04(0.04) | 0.00(0.02) | 0.00(0.03) |
| | Usually | 0.05(0.05) | 0.05(0.06) | 0.05(0.20) | 0.04(0.15) |
| Alcohol Use | Low | 0.08(0.06) | 0.08(0.07) | 0.19(0.35) | 0.15(0.28) |
| | Moderate | 0.05(0.05) | 0.05(0.05) | 0.01(0.09) | 0.02(0.10) |
| | Excessive | 0.11(0.09) | 0.12(0.10) | 0.40(0.46) | 0.43(0.42) |
| Recurrent Depression Status | Negative | 0.08(0.07) | 0.08(0.07) | 0.17(0.35) | 0.18(0.32) |
| | Positive | 0.05(0.05) | 0.04(0.05) | 0.07(0.22) | 0.06(0.18) |
| Economic Status | Low | 0.07(0.07) | 0.07(0.08) | 0.16(0.34) | 0.15(0.29) |
| | Middle | 0.04(0.04) | 0.03(0.04) | 0.00(0.04) | 0.00(0.03) |
| | High | 0.09(0.09) | 0.10(0.09) | 0.28(0.40) | 0.28(0.35) |

CAM-Score: Class Activation Map (CAM) based regional importance score. A higher CAM-Score indicates greater model attention and higher association with AD-related retinal features in that region, whereas a lower CAM-Score indicates lower contribution.

**Table 5. CAM-Scores Across Retinal Regions by Continuous Risk Factor Groups.**

| Risk Factors | CAM-Score; mean(std) | | | |
|---|---|---|---|---|
| | **Artery** | **Vein** | **Optic Disc** | **Optic Cup** |
| Age | 0.23(0.15) | 0.23(0.17) | 0.41(0.47) | 0.43(0.46) |
| Education | 0.25(0.16) | 0.25(0.18) | 0.24(0.39) | 0.31(0.39) |
| BMI | 0.23(0.16) | 0.23(0.17) | 0.37(0.46) | 0.41(0.45) |
| DBP | 0.21(0.17) | 0.21(0.18) | 0.39(0.48) | 0.40(0.47) |
| SBP | 0.23(0.16) | 0.23(0.17) | 0.37(0.46) | 0.40(0.46) |
| HbA1C | 0.22(0.15) | 0.23(0.17) | 0.35(0.45) | 0.39(0.45) |

CAM-Score: Class Activation Map (CAM) based regional importance score. A higher CAM-Score indicates greater model attention and higher association with AD-related retinal features in that region, whereas a lower CAM-Score indicates lower contribution.

BMI: Body Mass Index

DBP: Diastolic Blood Pressure

SBP: Systolic Blood Pressure

HbA1c: Hemoglobin A1c

*Saliency-Based Interpretability Reveals Anatomical Encoding of Risk Factors in CFP.*

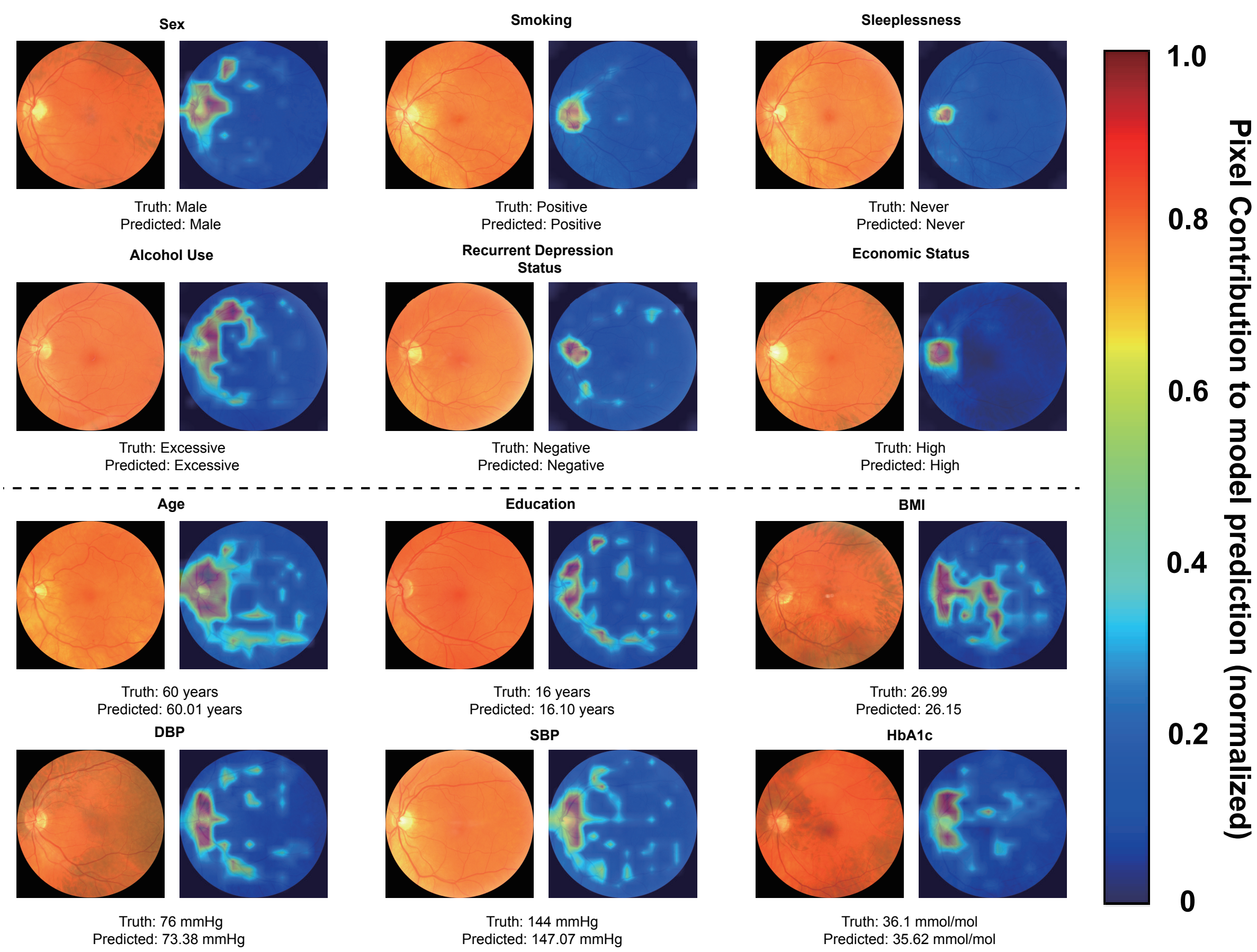

**Figure 1. ScoreCAM Saliency Maps for Predictions Across Alzheimer's Disease Risk Factors.** The top two rows display saliency maps for categorical risk factors, and the bottom rows depict saliency maps for continuous risk factors. The black horizontal line separates the categorical and continuous groups. For each visualization, the reference fundus image is shown alongside its corresponding maps. The heatmaps show pixel contribution to the model prediction. The region exhibiting higher intensity and spatial concentration demonstrate a higher model attribution, while regions with weaker signal indicate a lower contribution. Reproduced by kind permission of UK Biobank ©.

Figure 1 presents representative saliency maps for correct prediction cases across various risk factors, revealing consistent inference patterns in DL models to anatomically distinct retinal structures. The corresponding CAM-Scores acquired from the same saliency map visualization method are summarized in Tables 4 and 5 for both categorical and continuous risk factors. CAM-Score revealed distinct and interpretable structural patterns for several categorical variables at the population level. Based on the CAM-Score, the Sex prediction was strongly driven by the difference in optic disc and optic cup in males and females. Arterial and venous contributions were modest in sex classification, but consistently higher in males compared to females. This indicates all 4 regions were inferred by the model in sex classification. For Smoking, individuals with a history of smoking (Smoked) exhibited significantly elevated CAM-Scores in the optic disc and cup compared to individuals who never smoked (Never Smoked), suggesting that exposure to tobacco may be associated with differences in the optic nerve head regions detectable by the DL model. In Sleeplessness, the model focused more heavily on all 4 retinal structures in individuals who reported never experiencing sleeplessness, with substantially lower CAM-Scores in the people who reported experiencing sleeplessness sometimes, and slightly elevated but still modest scores in the individuals who had sleeplessness usually. This indicates that possible retinal structural changes are associated with sleep disruption. For Recurrent Depression Status, individuals without depressive symptoms had higher CAM-Scores in the optic disc and optic cup compared to the "Positive" group. Joint with the low contribution of vasculature, this implies that an association in the neuroretinal region is stronger with depressive symptoms. The non-linear pattern was observed in CAM-Scores for Alcohol Use and Economic Status. Individuals with “Excessive” alcohol consumption exhibited higher attribution across both vascular and optic cup & disc regions, but the “Moderate” group had the lowest CAM-Scores across all regions compared to the “Low” group. For Economic Status, the “High” income group showed the strongest structural saliency, followed by the “Low” income group.

For continuous risk factors, the model consistently exhibited high CAM-Scores across all four anatomical structures, indicating distributed retinal encoding of systemic characteristics. Age prediction was driven by contributions from both vascular and neuroretinal regions, reflecting age-associated changes in retinal morphology. Although the model’s predictive performance for Education was poor, CAM-Scores across all structures were unexpectedly observed, suggesting the presence of latent correlations cofounded by other AD-related variables. For BMI, high and consistent saliency was observed across artery, vein, disc, and cup, showing the link between obesity and retinal vascular and neuroretinal changes. Predictions for both blood pressures were similarly supported by strong saliency. DBP showed CAM-Scores of artery, vein, disc, and cup, while SBP yielded slightly higher scores across the same regions. These findings highlight that the model may capture retinal patterns related to the systemic hemodynamic changes. Scores for HbA1c were notable in the vascular and optic nerve regions.

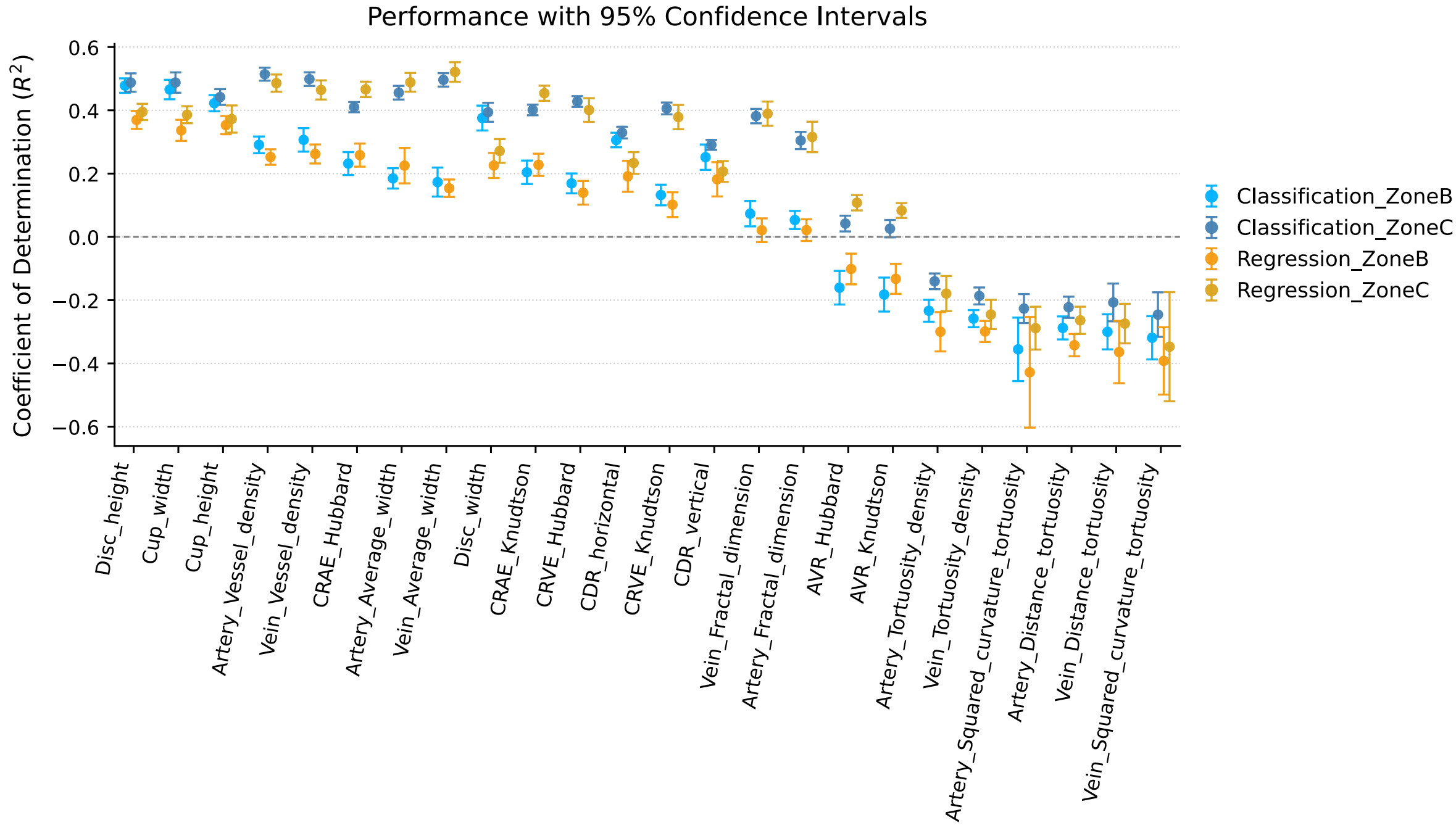

**Figure 2. Coefficient of Determination ($R^2$) of Retinal Morphology Features for Predicting AD Risk Factors Across Retinal Zones.** Bar plot showing the coefficient of determination ($R^2$) for each retinal morphological feature in predicting AD risk factors using DL features from Zone B (0.5–1.0 optic disc diameter) and Zone C (0.5–2.0 optic disc diameter). Both classification (categorical risk factors) and regression (continuous risk factors) tasks are represented. Blue and orange points represent features extracted from Zone B and Zone C, respectively, and error bars indicate 95% CIs computed across folds in 10-fold cross-validation. The horizontal black dashed line represents the $R^2 = 0$ threshold, below which the model performs worse than a null model predicting the mean.

*DL Features Accurately Encode Retinal Morphology*

The 10-fold cross-validation results of linear model prediction of retinal morphological features with DL based embeddings of colored fundus photography indicate the ability of the trained DL model in capturing retinal biomarkers, which further provides a reliability of its prediction (Figure 2).

*Optic Disc and Cup Morphology.* High and consistent predictive performance was observed for structural parameters of the optic nerve head, including disc height, disc width, cup height, and cup width. Notably, disc height achieved a strong $R^2$ across classification (Zone C:

0.49 ± 0.01; Zone B: 0.48 ± 0.01) and regression models (Zone C: 0.40 ± 0.02; Zone B: 0.38 ± 0.01). Similar patterns were observed for disc width (Classification-Zone C: $R^2$=0.39±0.01; Classification-Zone B: $R^2$=0.37±0.01; Regression-Zone C: $R^2$=0.28±0.01; Regression-Zone B: $R^2$=0.22±0.01), cup height (Classification-Zone C: $R^2$=0.44±0.02; Classification-Zone B: $R^2$=0.42±0.02; Regression-Zone C: $R^2$=0.37±0.02; Regression-Zone B: $R^2$=0.36±0.01), and cup width (Classification-Zone C: $R^2$=0.49±0.01; Classification-Zone B: $R^2$=0.46±0.01; Regression-Zone C: $R^2$=0.39±0.01; Regression-Zone B: $R^2$=0.35±0.01), indicating the model's ability of encoding of optic nerve head morphology. Limited, but above-chance predictive performance was observed for vertical and horizontal cup-to-disc ratios (CDRs), further supporting the DL model's capacity to capture clinically relevant optic features during prediction.

*Vascular Feature Metrics.* The varying predictive performance was observed for retinal vascular features, particularly within Zone C. Fractal dimension (Classification-Zone C: $R^2$=0.31±0.01; Classification-Zone B: $R^2$=0.05±0.01; Regression-Zone C: $R^2$=0.31±0.02; Regression-Zone B: $R^2$=0.01±0.02), vessel density (Classification-Zone C: $R^2$=0.52±0.01; Classification-Zone B: $R^2$=0.28±0.01; Regression-Zone C: $R^2$=0.48±0.02; Regression-Zone B: $R^2$=0.25±0.02), and average width (Classification-Zone C: $R^2$=0.45±0.01; Classification-Zone B: $R^2$=0.19±0.01; Regression-Zone C: $R^2$=0.49±0.01; Regression-Zone B: $R^2$=0.21±0.02) showed strong associations with DL features. Similar trends were observed for venous features, including fractal dimension (Classification-Zone C: $R^2$=0.37±0.01; Classification-Zone B: $R^2$=0.08±0.02; Regression-Zone C: $R^2$=0.39±0.01; Regression-Zone B: $R^2$=0.01±0.01), vessel density (Classification-Zone C: $R^2$=0.50±0.01; Classification-Zone B: $R^2$=0.31±0.01; Regression-Zone C: $R^2$=0.46±0.02; Regression-Zone B: $R^2$=0.26±0.02), and average width (Classification-Zone C: $R^2$=0.49±0.01; Classification-Zone B: $R^2$=0.18±0.01; Regression-Zone C: $R^2$=0.52±0.01; Regression-Zone B: $R^2$=0.16±0.01). These findings underscore the DL model's ability to encode vascular architecture network complexity, particularly in the peripapillary region.

*High-Order Tortuosity Metrics.* In contrast to vascular feature metrics, higher-order features such as arterial and venous tortuosity (e.g., squared curvature tortuosity, tortuosity density) were poorly predicted by the linear model, with several $R^2$ below zero. For instance, arterial squared curvature tortuosity in Zone B showed $R^2 = -0.49 \pm 0.09$ (regression) and $-0.36 \pm 0.04$ (classification), suggesting the limitations of DL representation in curvature-based morphological characteristics from the DL feature space.

*Vessel Caliber Metrics.* Both CRAE and CRVE, computed with Hubbard and Knudtson formulas, were predicted above chance in Zone C. For instance, CRAE-Knudtson achieved $R^2$ of $0.41 \pm 0.01$ (classification) and $0.46 \pm 0.01$ (regression), while CRVE-Knudtson yielded $0.40 \pm 0.01$ and $0.38 \pm 0.01$. Comparable results were observed with the prediction results of the Hubbard method-based metric. These outcomes suggest that the DL model captures vessel caliber information despite not being explicitly trained. In contrast, the arteriovenous ratio (AVR) measures exhibited only mild predictability in Zone C and poor performance in Zone B.

*Deep Learning Features Mirror True Biological Variation*

The results of the statistical analysis between retinal morphology measures and risk factors indicate the biologically relevant variations implicitly captured by the DL model. Supplementary Tables 5-10 show the results of independent two-sample t-tests to compare the means of retinal morphology measures between classes from each risk factor. Supplementary Tables 11-16 show the result of Pearson's correlation coefficients (R) to quantify linear associations between retinal morphology features and risk factors. The significant biological variations related to risk factors are described.

*Sex.* No significant difference in sex was found in optic cup or cup-to-disc ratio (CDR). However, males exhibited a larger optic disc width compared to females. In both Zone B and C, the arterial width of males was lower compared to that of females. In venous features, the fractal dimension was higher in males, while the average width was lower. In addition, males showed lower vein squared curvature tortuosity in Zone C, and less distance tortuosity in both zones. For caliber measurements, CRAE with both methods (Hubbard and Knudtson) was lower in males, while CRVE was higher in males across both zones, except CRVE in Zone B. AVR was consistently lower in males, indicating proportionally narrower arteries in males.

*Smoking.* Except smaller disc height among smokers compared to individuals who never smoked, smoking status showed limited influence on optic disc and cup features. Smokers had a higher average arterial width in both Zone B and C. This trend was aligned in venous width in both zones. In addition, lower venous density at Zone B among individuals who never smoked was observed. For caliber measurements, CRAE and CRVE (Hubbard) were lower in never smokers than in smokers in both Zone B and Zone C. AVR was higher in never smokers in both zones. These trends were consistent with the Knudtson method.

*Sleeplessness.* Individuals reporting sleeplessness had lower disc height and fractal dimensions of the artery and vein. In addition, the lower venous density and wider venous width were observed with increased sleeplessness in both zones. For vessel calibers, AVR (Hubbard) increased with increasing sleeplessness in Zone B, reflecting proportionally wider veins or narrower arteries, and similar trends were found in AVR(Knudtson).

*Alcohol Use.* Individuals with higher alcohol consumption reported smaller disc height. In arteries, the narrower arteries and vessel density were observed to be proportional to the alcohol consumption in both Zone B and Zone C. The fractal dimension was highest in the moderate group, following the inverted U shape. The venous vasculature of heavy drinkers showed lower vessel density, distance tortuosity, and squared curvature tortuosity, suggesting a smoother and less tortuous network. For vessel calibers, CRAE was smaller in the moderate and excessive drinkers compared to low drinkers in both zones, consistent across both methods (Hubbard and Knudtson). In contrast, CRVE was only lower in Zone C, and AVR was inversely proportional to alcohol consumption using both methods and in both zones.

*Recurrent Depression Status.* Participants with recurrent depression exhibited smaller disc and cup and lower horizontal CDR compared with non-depressed individuals. While no difference was observed in arterial measures except arterial width, wider venous width and lower fractal dimension were observed, suggesting vascular dilation and reduced vasculature complexity. Other measures were largely unchanged.

*Economic Status.* Individuals with high income showed only the lower disc heights. For arterial features, the fractal dimension was larger in high-income individuals in Zone C, suggesting more complex vascular branching in wealthier individuals. In addition, the average width decreased with the increase in economic status. For venous fractal dimension, the relationship was proportional to the rising economic status in both zones, indicating more complex venous networks in higher socioeconomic groups. Average venous width was narrower in the high economic status group across both zones, showing a consistent narrowing trend. Notably, venous tortuosity density was significantly lower in the high economic group, reflecting smoother venous paths.

*Age.* Negative correlation between age and disc height and horizontal CDR was found. Moreover, the inverse correlation between arterial and venous fractal dimension and vessel density was present, reflecting the vascular simplification in older populations. Arteries became wider and veins wider with age increase. The venous tortuosity showed an increase with age.

All vessel caliber measures, regardless of methods, had significant negative correlations with age in both zones.

*Education.* No significant correlations were observed in disc or cup dimensions with education. Higher education yielded greater fractal complexity and narrower width for both arteries and veins. In artery-specific, distance tortuosity also showed a positive correlation in Zone C. For caliber measurements, CRVE of both methods was negatively correlated with education in Zone C, suggesting venular narrowing in more educated individuals.

*BMI.* Higher BMI is linked inversely with average width, vessel density, distance tortuosity, squared curvature tortuosity, and fractal dimension. In contrast, for veins, the average width, vessel density, and fractal dimension had positive correlations, which were asymmetric arterial and venous networks with respect to BMI levels. For caliber measurements, all CRAE and AVR declined with BMI in both zones. Conversely, CRVE increased with BMI.

*DBP & SBP.* Blood pressure measures were positively associated with disc width, suggesting mild disc expansion with higher blood pressure. Arterial fractal dimension, vessel density, and average width were negatively correlated with DBP in both zones. Artery squared curvature tortuosity showed a slight positive correlation in Zone C. In the venous network, tortuosity density in Zone C showed a small but significant positive correlation with DBP, suggesting a subtle venular response. Fractal dimension and vessel density were negatively correlated with SBP in both zones. In vessel caliber, CRAE, and AVR exhibited a strong negative association with both blood pressure levels and CRVE for only SBP.

*HbA1c*. Higher HbA1c levels were associated with reduced complexity and density, with wider and more tortuous veins. In arteries, fractal dimension and vessel density were negatively associated with HbA1c, while average width showed an opposite trend. Zone B arterial metrics showed no significant relationships, except for a mild negative correlation in artery distance tortuosity. In veins, average width and tortuosity density showed positive correlation, while fractal dimension and vessel density were negatively correlated. The measures in Zone B also showed a significant positive correlation for vein average width and tortuosity density. For vessel caliber, negative correlations were observed with CRAE and AVR.

**Table 6. Mann–Whitney U Test Results for CAM-Scores of Categorical Risk Factors: Incident AD Compared with Matched Controls.**

| Risk Factors | Class | Median (Interquartile Range; Q1-Q3) | | | | | | | |
|---|---|---|---|---|---|---|---|---|---|
| | | Artery | | Vein | | Optic Disc | | Optic Cup | |
| | | NC | AD | NC | AD | NC | AD | NC | AD |
| Sex | Female | 0.08 (0.05–0.15) | 0.10 (0.07–0.16) | 0.08 (0.04–0.14) | 0.08 (0.05–0.11) | 0.16 (0.10–0.25) | 0.26 (0.26–0.26) | 0.47 (0.24–0.70) | 0.10 (0.10–0.10) |
| | Male | 0.14 (0.09–0.19) | 0.14 (0.10–0.21) | 0.14 (0.10–0.19) | 0.16 (0.11–0.23) | 0.99 (0.85–1.00) | 1.00 (0.84–1.00) | 1.00 (1.00–1.00) | 1.00 (0.97–1.00) |
| Smoking | Never Smoked | **0.06 (0.04–0.12)** | **0.05 (0.02–0.09)** | **0.07 (0.03–0.12)** | **0.04 (0.02–0.08)** | **0.05 (0.01–0.12)** | **0.37 (0.25–0.62)** | 0.58 (0.46–0.70) | 0.75 (0.46–1.00) |
| | Smoked | 0.08 (0.06–0.13) | 0.09 (0.06–0.12) | 0.09 (0.07–0.12) | 0.10 (0.07–0.14) | 0.95 (0.71–1.00) | 0.95 (0.78–1.00) | 1.00 (1.00–1.00) | 1.00 (1.00–1.00) |
| Sleeplessness | Never | **0.05 (0.02–0.13)** | **0.09 (0.04–0.16)** | **0.05 (0.02–0.13)** | **0.09 (0.04–0.17)** | **0.51 (0.32–0.77)** | **0.81 (0.37–0.97)** | 0.69 (0.30–1.00) | 0.99 (0.37–1.00) |
| | Sometimes | 0.03 (0.02–0.06) | 0.04 (0.01–0.06) | 0.04 (0.02–0.07) | 0.02 (0.01–0.06) | 0.13 (0.08–0.24) | 0.23 (0.23–0.23) | 0.63 (0.63–0.63) | 0.05 (0.03–0.07) |
| | Usually | 0.03 (0.02–0.08) | 0.04 (0.01–0.06) | **0.04 (0.01–0.10)** | **0.03 (0.01–0.06)** | 0.04 (0.02–0.13) | 0.13 (0.09–0.34) | 0.68 (0.46–0.81) | 0.56 (0.08–0.98) |
| Alcohol Use | Low | 0.06 (0.03–0.11) | 0.05 (0.03–0.09) | 0.05 (0.02–0.10) | 0.04 (0.02–0.09) | 0.25 (0.01–0.55) | 0.36 (0.12–0.55) | 0.88 (0.31–1.00) | 0.56 (0.08–0.85) |
| | Moderate | 0.03 (0.01–0.05) | 0.03 (0.01–0.06) | 0.02 (0.01–0.06) | 0.03 (0.01–0.06) | 0.10 (0.06–0.32) | 0.20 (0.11–0.65) | 0.45 (0.40–0.50) | 0.89 (0.45–0.94) |
| | Excessive | 0.11 (0.06–0.17) | 0.14 (0.08–0.21) | **0.12 (0.05–0.18)** | **0.16 (0.07–0.22)** | 0.80 (0.58–1.00) | 0.87 (0.40–1.00) | 1.00 (0.67–1.00) | 1.00 (0.40–1.00) |
| Recurrent Depression Status | Negative | 0.06 (0.02–0.11) | 0.05 (0.03–0.12) | 0.08 (0.03–0.14) | 0.09 (0.03–0.14) | 0.43 (0.13–0.69) | 0.57 (0.20–0.91) | 0.80 (0.18–0.98) | 0.84 (0.28–1.00) |
| | Positive | 0.02 (0.01–0.05) | 0.03 (0.01–0.06) | **0.01 (0.01–0.04)** | **0.02 (0.01–0.05)** | 0.24 (0.11–0.38) | 0.39 (0.13–0.72) | **0.49 (0.14–0.74)** | **0.91 (0.46–0.95)** |
| Economic Status | Low | **0.07 (0.03–0.14)** | **0.04 (0.02–0.09)** | **0.08 (0.02–0.16)** | **0.03 (0.01–0.09)** | 0.43 (0.19–0.80) | 0.61 (0.18–0.73) | 0.56 (0.31–1.00) | 0.76 (0.40–0.99) |
| | Middle | **0.02 (0.01–0.04)** | **0.04 (0.03–0.06)** | 0.02 (0.01–0.04) | 0.03 (0.01–0.06) | 0.00 (0.00–0.00) | 0.01 (0.01–0.01) | N/A | N/A |
| | High | **0.06 (0.02–0.11)** | **0.09 (0.04–0.16)** | **0.06 (0.02–0.11)** | **0.10 (0.04–0.19)** | 0.38 (0.21–0.76) | 0.63 (0.19–0.94) | 0.59 (0.14–1.00) | 0.99 (0.17–1.00) |

* p-value <0.05 is bolded. The p-values were calculated based on non-zero values. Non-zero rate information is available in Supplementary Table 17.

**Table 7. Mann–Whitney U Test Results for CAM-Scores of Continuous Risk Factors: Incident AD Compared with Matched Controls.**

| Risk Factors | Median (Interquartile Range; Q1-Q3) | | | | | | | |
|---|---|---|---|---|---|---|---|---|
| | Artery | | Vein | | Optic Disc | | Optic Cup | |
| | NC | AD | NC | AD | NC | AD | NC | AD |
| Age | 0.15 (0.08–0.34) | 0.15 (0.09–0.32) | 0.12 (0.06–0.37) | 0.13 (0.06–0.40) | 0.84 (0.52–1.00) | 0.98 (0.67–1.00) | 1.00 (0.81–1.00) | 1.00 (0.68–1.00) |
| Education | 0.24 (0.08–0.41) | 0.30 (0.08–0.42) | 0.27 (0.07–0.45) | 0.34 (0.07–0.45) | **0.61 (0.27–0.97)** | **0.91 (0.73–1.00)** | **0.70 (0.32–1.00)** | **1.00 (0.76–1.00)** |
| BMI | 0.14 (0.06–0.39) | 0.26 (0.07–0.41) | 0.15 (0.05–0.40) | 0.26 (0.05–0.43) | 0.98 (0.91–1.00) | 1.00 (0.80–1.00) | 1.00 (0.98–1.00) | 1.00 (0.88–1.00) |
| DBP | 0.12 (0.07–0.40) | 0.10 (0.06–0.37) | 0.10 (0.04–0.45) | 0.11 (0.04–0.39) | 1.00 (1.00–1.00) | 1.00 (0.94–1.00) | 1.00 (1.00–1.00) | 1.00 (1.00–1.00) |
| SBP | 0.26 (0.08–0.42) | 0.31 (0.07–0.43) | 0.23 (0.07–0.44) | 0.31 (0.07–0.48) | 0.99 (0.61–1.00) | 0.98 (0.83–1.00) | 1.00 (0.62–1.00) | 1.00 (0.85–1.00) |
| HbA1c | **0.23 (0.09–0.40)** | **0.12 (0.06–0.35)** | 0.20 (0.06–0.40) | 0.11 (0.05–0.38) | 0.91 (0.76–1.00) | 0.97 (0.72–1.00) | 1.00 (0.69–1.00) | 1.00 (0.77–1.00) |

* p-value <0.05 is bolded. The p-values were calculated based on non-zero values. Non-zero rate information is available in Supplementary Table 18.

*Saliency-Derived Scores from Risk Factors Overlap with Retinal Structural Differences of Incident Alzheimer's Disease*

In the previous section, the DL-derived features from CFP effectively encoded biologically meaningful retinal structural information. Building on this observation, DL-derived features were further validated to reveal that the structural alterations associated with AD-related risk factors can overlap with future AD onset. Therefore, the Mann-Whitney U test CAM-Scores was performed between incident AD subjects and selected controls (Demographic details between incident AD and control are described in Supplementary Table 2).

For CAM-Scores of categorical risk factors (Table 6), incident AD subjects showed significant differences in at least one retinal structure. For smoking status, sleeplessness, and economic status, they displayed significant alternations across all four retinal structures, suggesting that the model may capture the multi-structural retinal signatures distinguishing individuals who will later develop AD. In contrast, for continuous variables (Table 7), CAM-Scores showed limited ability to differentiate incident AD cases from controls. The CAM-Score for the artery in HbA1c prediction was significantly different between incident AD and controls. Importantly, the goal of this study is not to position CFP-based DL models as clinical diagnostic tools for AD, but rather to determine whether the retina contains latent structural signatures of upstream systemic factors known to influence AD vulnerability. This distinction reinforces the clinical relevance of our findings, demonstrating that systemic risks reside in retinal morphology and are detectable by DL models.

## Discussion

This study demonstrates that DL models show heterogeneous performance across AD-related risk factor prediction, yet consistently reveal retinal correlates that align with established biological patterns even in domains with limited or weak predictive performance. The strongest predictive signals were observed for age, sex, and followed by limited but above chance performance for blood pressure and smoking status. Notably, risk factors that were not previously well characterized in the context of CFP (e.g., sleeplessness, economic status) showed moderate or weak predictive performance accompanied by localized saliency in retinal structures, suggesting that the model may be detecting structural patterns warranting further investigation. Across risk domains, CFP-derived feature representations captured anatomically coherent structures and retinal morphometrics through both saliency analysis and linear prediction of morphological measures, with the exception of tortuosity measures. Finally, structural differences between incident AD cases and matched controls were reflected in the saliency-based scoring metric, indicating that retinal patterns associated with upstream modifiable risk factors partially overlap with AD vulnerability before clinical onset.

The outperformance of the DL models compared to the morphometry-based machine learning baseline indicates their capacity to learn additional retinal structural patterns beyond predefined morphometric descriptors. The DL model's performance trend is consistent with prior studies that employed DL for risk factor modeling with CFPs [22–24]. However, existing work has not directly compared CFP-derived feature-based approaches, leaving the specific advantage of DL for capturing structural associations between CFPs and risk factors insufficiently validated. Notably, the findings of risk factors with low predictive performance should be interpreted as hypothesis-generating, rather than evidence of robust association. Despite the limited performance, considering that economic status, alcohol use, and sleeplessness have not yet been seriously tackled in the retinal domain, the observed predictive power of DL models underscores the need for deeper investigation in this domain to characterize the relationships between these factors and structural correlates of the retina.

The model's saliency patterns and downstream associations aligned with known retinal variations across these risk domains. Sex differences in vascular and optic nerve morphology are well established in our analyses. The larger disc widths, narrower arteries, wider venules, and reduced venous tortuosity in males are plausible with males' higher vascular stiffness and metabolic load [47]. For smoking, it is reported that endothelial dysfunction and microvascular dilation may present [48,49]. The larger vessel calibers in smokers for both arteries and veins from our study align with vascular remodeling and inflammation known to affect retinal microvasculature [50]. For age, the most representative risk factors of AD and related dementia, results demonstrated vascular stiffening, lower density, and optic nerve head thinning. This is also well-established in age-related remodeling of retinal structures [47,51]. Obesity is known to drive the inflammatory response and endothelial dysfunction, producing asymmetric arterial and venous changes. Our results in asymmetric arterial-venous changes at higher BMI reflect the known patterns [52]. As shown in previous hypertension research demonstrating the narrowed vessel, reduced branches, and complexity, the negative correlations with fractal dimension, density, and calibers mirror the hypertensive retinopathy features [53]. For HbA1c, results follow the previous findings from hyperglycemia, leading to capillary dropout and venous dilation [54].

While the influence of major cardiovascular and systemic risk factors mentioned above is well studied, our findings highlight retinal correlates of underexplored factors, including psychiatric symptoms, socioeconomic status, alcohol consumption, and sleep quality in CFP modalities. The major recurrent depression showed one of the consistent associations, with smaller disc and cup dimensions and lower horizontal CDR in depression positive individuals. This pattern is plausible given vascular and neuroinflammatory changes observed in depressive disorders, although existing work primarily focused on OCT-based markers [55,56], while a few studies were looking into CFP for mental disorders [57]. The retinal biomarkers specific to depression and related psychiatric disease are still understudied [58,59], but our findings suggest that the structural variations in neuroretinal regions (the optic disc and cup) can be a potential source. While the predictive performance of the economic status of our model was limited, it also demonstrated measurable associations with higher income individuals, showing larger disc size and greater arterial and venous fractal dimensions. Prior work has reported a moderate relationship between socioeconomic status and retinal vascular characteristics in disease-specific populations [60–62], while our findings are derived from the normal population. Socioeconomic risk factors are likely mediated or influence the behavioral and environmental pathways, which are difficult to be directly captured by a single biological structure. Therefore, the observed associations between CFP features and economic status require further investigation to discover indirect correlates. The alcohol use exhibited modest predictive performance, indicating the CFP features alone may provide limited discriminatory information. However, our saliency analysis suggested potential structural patterns, including smaller disc height and vasculature alternations for the population with high alcohol consumption. It is compatible with known microvascular changes associated with chronic alcohol exposure, although most prior evidence is derived from RNFL, rather than structures visible from CFP [63]. Lastly, for sleeplessness, the predictive performance of the model was relatively low, suggesting the limited sensitivity of the direct relationship between retinal and sleep-related systemic effects. The result aligned with narrower venular calibers from frequent sleeplessness in the analysis, but the result also presents a correlation with reduced vascular complexity. However, under the consideration that very limited research on the association between sleep disorders and retinal image has been performed, our findings can be interpreted as preliminary and require validation in specifically designed studies for sleep-related retinal changes [64]. Importantly, the interpretation of the association between retinal morphology and AD risk factors with low predictive performance should remain indirect and exploratory. Despite limitations, this line of findings supports the biological plausibility of retinal correlates derived from CFP for a range of risk factors discovered by DL while underscoring the need for further investigation to clarify their specificity and underlying mechanisms.

Based on both qualitative visualization and quantitative results with a saliency-based scoring metric, the DL model prediction of risk factors referred to the primary retinal structures. Notably, our quantitative result assures the reference is at the group level, not cherry picking the good results. Moreover, CFP features encoded by the DL model also robustly capture anatomical structures, particularly the optic disc, cup, and vascular caliber, while poorly encoding tortuosity metrics, which were not available with only saliency-based qualitative analysis. While prior studies of applying CFP to DL models have commonly applied Grad-CAM to visually confirm the model's interpretability, showing its inference on anatomically plausible retinal regions, few have systematically assessed the extent to which these models encode quantifiable morphological features at the population level. In this study, we extend interpretability by introducing CAM-Scores and DL feature regression analyses to evaluate how well structural and vascular characteristics are captured by the model. The findings indicate that the DL representations robustly encode optic nerve head morphology and vessel calibers, while exhibiting limited sensitivity to high-order tortuosity and AVR. These lines of result are meaningful in further clinical and engineering studies, as several risk factors with moderate or poor predictive performance demonstrated significant associations with tortuosity measures and AVR (alcohol use, economic status, blood pressure, education, BMI, and HbA1c). The result suggests that future risk prediction models with the establishment of complex retinal vascular geometry representations may improve their prediction capability of systemic risk factors.

The exploratory retrospective analysis showed a significant difference between cognitive normal controls and the incident AD group. Thus, despite our findings based on DL being exploratory due to the small number of incident AD group, these associations between DL features and classical retinal measurements suggest that CFP-derived representations from the risk factor prediction model may reflect biologically relevant variations of AD vulnerability, providing interpretability of model predictions and reliability for clinical implications. A growing body of research demonstrates the link between retinal structures and neurodegenerative processes in AD, particularly the optic nerve head morphology. Prior studies have demonstrated that AD is associated with enlarged CDR and reduced rim area [65]. Similar observation was found in prodromal population, where abnormally large CDR correlated with lower total brain volume specific to frontal and occipital lobes, supporting the interpretation that optic nerve head morphology demonstrates the aspects of brain health. Retinal vascular features, including reduced fractal dimensions [66], sparser microvasculature in AD patients [67], and characteristic patterns of narrower arterioles and wider venules linked to dementia [68] and AD [69,70] specifically support the neurodegenerative association with retinal morphology found in CFP. However, many of these same retinal alternations are also strongly influenced by risk factors, highlighting that DL features extracted from CFP may encode general structural burden rather than pathologically unique features to AD. The CAM-Score comparison between cognitively normal individuals and those who later developed AD shows the capability of DL-derived features in capturing retinal structural changes related to AD.

This study has several limitations. First, the cohort in this cross-sectional study predominantly comprised Caucasian individuals (~94%), limiting the generalizability of our findings to other ethnic populations. Prior studies have demonstrated notable differences in retinal anatomy across ethnicities; for instance, African and Asian populations typically exhibit larger optic discs and wider vessel calibers compared to Caucasians [71,72]. Such variations may affect both model performance and the interpretation of associations between retinal morphology and risk factors. Validation in a more ethnically diverse cohort is therefore essential. Second, our analyses were cross-sectional and relied primarily on linear associations and group-level comparisons, which are limited to capturing the complex and potentially nonlinear relationships among socioeconomic, behavioral, and metabolic factors. Confounding and interaction effects were not modeled explicitly, and we cannot infer temporal relationships between systemic risks and retinal structural changes contributing to AD-specific biomarkers. Therefore, longitudinal studies will be required to determine whether the patterns found by the DL model precede, follow, or simply coincide with systemic health changes relevant to AD risk. Third, although CAM-Scores improve interpretability, they reveal only where the model attributes importance, not reflecting the underlying mechanistic links. The difference in CAM-Scores between incident AD and control groups may reflect shared systemic pathways- such as vascular or metabolic burden, rather than early AD-specific changes. While the model captures retinal structural differences correlated with incident AD, these should still be interpreted as AD-related but not AD-specific. In addition, the predominance of significance among categorical risk factors, with HbA1c being the only significant continuous predictor, may have derived from the methodological limitation of CAM. Otherwise, it suggests that DL-encoded structural differences may be more sensitive to risk stratification boundaries rather than fine-grained physiological variations. The limited cohort size for comparison also limits the statistical power of the result in incident AD analysis. Therefore, larger cohorts and granular modeling may be needed to detect subtler continuous associations. Moreover, this study did not explicitly accounted for the paired nature of left and right eyes from the same participant. The left and right eye does not share ideal morphometry, but not independent. Treating eyes as independent observations may underestimate the variance due to within-subject correlation. However, the analysis was intended at the eye-level, consistent with the formulation of the classification and regression task of our study. Given the symmetry of ocular structure and our focus on eye-specific rediction rather than subject level inference, this design choice is not expected to affect primary conclusion of this study [73–75]. Finally, our study does not claim to introduce a new predictive architecture. Rather, our contribution lies in finding relationships of understudied AD-related risk factors and CFP, and demonstrating how an existing DL model captures biologically interpretable retinal variations associated with AD-related risk domains. However, based on the linear probing result, the lack of tortuosity prediction in our linear probing analysis suggests the potential target of architecture or algorithm improvement for better risk factor prediction performance.

In this study, DL models trained on CFP were able to predict a broad range of AD-related risk factors, including demographic, behavioral, metabolic, and psychiatric variables, with varying degrees of performance. Strong performance for age and sex aligns with well-established biological determinants of the retina. In contrast, for several other factors including depression, alcohol use, and economic status demonstrated moderate or weak predictive performance. The prediction results need cautious interpretation. While the AUROC values of factors including alcohol consumption, economic status, and sleeplessness are above chance level, the relatively limited predictive accuracy indicates the retinal features may contain only weak or indirect predictive signals related to systemic status. Interpretability analysis using CAM-Scores and regression with retinal morphology features showed that the model robustly captured anatomical information related to the optic disc, cup, and vascular calibers, although sensitivity to high-order tortuosity metrics remained limited. Moreover, the CAM-Scores comparison between incident AD groups and controls demonstrated that the DL saliency of CFP prediction risk factors may detect the retinal structural differences related to both AD and risks, which holds potential in risk stratification of the population at risk of AD. The convergence between predictive features and statistical associations reinforces the biological plausibility of the DL's outputs and reinforces retinal imaging as a promising non-invasive method for examining systemic health changes relevant to AD risk. Our integrative design provides a unified analytical pipeline that bridges predictive performance with mechanistic interpretability, a connection not established in prior work on CFP-based dementia or risk-factor modeling. Our work demonstrates that DL models can extract retinal structural patterns aligned with systemic factors contributing to AD vulnerability. By linking model representations to both classical morphometrics and incident AD outcomes, the study provides a foundation for future work to explore whether CFP-derived features may serve as scalable biomarkers of systemic health domains relevant to neurodegeneration. In future studies, the longitudinal validation, including the larger AD-specific cohort, is required to establish causal and specific relevance.

## Acknowledgements

This research has been conducted using the UK Biobank Resource under Application Number

48388.

## Author Contributions

Seowung Leem (Data curation; Formal analysis; Investivation; Methodology; Validation; Visualization; Writing – original draft); Yunchao Yang (Methodology; Resources, Software; Validation; Writing – review & editing); Adam Woods (Conceptualization; Methodology; Resources; Supervision; Validation; Writing – review & editing); Ruogu Fang (Conceptualization; Funding acquisition; Project administration; Resources; Supervision; Validation; Writing – review & editing);

## Ethical Considerations

Our analysis was a cross-sectional study based on the secondary analysis of de-identified data under application 48388. The original data was approved by the National Research Ethics Service North West–Haydock Research Ethics Committee.

## Consent to Participate

Participants provided informed consent and approval from the National Research Ethics Service North West–Haydock Research Ethics Committee.

## Consent to Publication

Not applicable.

## Declaration of Conflicting Interests:

The authors declare that there is no conflict of interest to report.

## Funding

This material is based upon work supported by the National Science Foundation under Grant No. (NSF 2123809).

## Data Availability Statement

This research has been conducted using data from the UK Biobank under application 48388. The application of the de-identified data is available in the UK Biobank (https://www.ukbiobank.ac.uk/). The code for analysis is provided in https://github.com/lab-smile/DeepRetinaADRisk.